\documentclass[10pt, conference, compsocconf]{IEEEtran}
\usepackage{cite}
\usepackage{tabularx,graphicx}
\usepackage{amsmath,amssymb,amsfonts}
\usepackage{algorithmic}
\usepackage{textcomp}
\usepackage{comment}
\usepackage{float}
\usepackage{stfloats}
\usepackage{array}
\usepackage{booktabs} 
\usepackage{csquotes}
\usepackage{adjustbox}
\usepackage{threeparttable}
\usepackage{url}
\usepackage{rotating}
\usepackage{multirow}
\usepackage{subcaption}
\newcommand{\etal}{\textit{et al}. }

\newcommand\blfootnote[1]{%
	\begingroup
	\renewcommand\thefootnote{}\footnote{#1}%
	\addtocounter{footnote}{-1}%
	\endgroup
}

\begin{document}
	%
	\title{ICDAR2019 Robust Reading Challenge on Arbitrary-Shaped Text - RRC-ArT}

	
	\author{\IEEEauthorblockN{Chee-Kheng Chng$^{*1}$, Yuliang Liu$^{*2}$, Yipeng Sun$^{*3}$, Chun Chet Ng$^{1}$, Canjie Luo$^{2}$, Zihan Ni$^{3}$,\\ ChuanMing Fang$^{2}$, Shuaitao Zhang$^{2}$, Junyu Han$^{3}$, Errui Ding$^{3}$, Jingtuo Liu$^{3}$,\\ Dimosthenis Karatzas$^{4}$, Chee Seng Chan$^{1}$, Lianwen Jin$^{2}$}
		\IEEEauthorblockA{$^{1}$Fac. of Comp. Sci. \& Info. Tech., University of Malaya, Malaysia\\
			$^{2}$South China University of Technology\\
			$^{3}$Baidu Inc, Beijing, China\\
			$^{4}$Computer Vision Center, Universitat Aut\`onoma de Barcelona, Spain\\
	}}
	
	
	%


	\maketitle

	\begin{abstract}
		This paper reports the ICDAR2019 Robust Reading Challenge on Arbitrary-Shaped Text - \textit{RRC-ArT} that consists of three major challenges: i) scene text detection, ii) scene text recognition, and iii) scene text spotting. A total of 78 submissions from 46 unique teams/individuals were received for this competition. The top performing score of each challenge is as follows: i) T1 - 82.65\%, ii) T2.1 - 74.3\%, iii) T2.2 - 85.32\%, iv) T3.1 - 53.86\%, and v) T3.2 - 54.91\%. Apart from the results, this paper also details the \textit{ArT} dataset, tasks description, evaluation metrics and participants' methods. The dataset, the evaluation kit as  well as the results are publicly available at the challenge website\footnote{https://rrc.cvc.uab.es/?ch=14}.
	\end{abstract}
	

	%
	\IEEEpeerreviewmaketitle

	\section{Introduction}\label{introduction}
	Text in the wild comes in a variety of shapes. However, linear text arrangements, be it horizontal or rotated (as defined by multi-oriented text datasets like ICDAR2015\cite{karatzas2015icdar} and MSRA-TD500\cite{yao2012detecting}) dominate existing popular datasets such as ICDAR2013 \cite{karatzas2013icdar}, ICDAR2015 \cite{karatzas2015icdar}, COCO-Text \cite{veit2016cocotext}. Text instances arranged in curved or other irregular arrangements, as pointed out in Total-Text \cite{Chng2017TotalTextAC} and SCUT-CTW1500 \cite{yuliang2017detecting}, despite their commonness in our real world scenes, are rarely seen in the mentioned datasets. As a result, text detection models properly considering arbitrary-shaped text are relatively uncommon. In addition, recent studies \cite{Chng2017TotalTextAC,yuliang2017detecting, long2018textsnake, lyu2018mask, long2018scene} point out that existing state-of-the-art scene text detection models perform poorly against such data. Their studies suggest that a major design change is needed to handle the wild nature of arbitrary-shaped text instances.\blfootnote{*Authors contributed equally}
	
	Motivated by \cite{Chng2017TotalTextAC} and \cite{yuliang2017detecting}, numerous research works \cite{long2018textsnake, lyu2018mask} have demonstrated their interest in tackling the curved text reading problem. These studies suggest that some principle design changes are necessary in order to produce a tight polygon detection result, which is capable of binding arbitrary-shaped texts tightly. One example is the increment of the regression variables in order to cater for the higher count of vertices of a curved text region \cite{yuliang2017detecting}. Meanwhile, \cite{lyu2018mask} took advantage of the segmentation-based approach to address this problem. However, since the testing sets of \cite{Chng2017TotalTextAC} and \cite{yuliang2017detecting} consist of only 300 and 500 images, respectively, it is hard to draw conclusive claims based on them due to its relatively small sample size. Hence, we combined all the released images and ground truth in both of the mentioned datasets as the training set for this competition, and at the same time collected new images with similar attributes (i.e. high existence of arbitrary-shaped text alongside horizontal and multi-oriented text) to increase the size of both the training and testing set.
	
	This competition is a natural extension to all the previous RRC competitions, and consists of three main tasks: i) scene text detection, ii) scene text recognition, and iii) scene text spotting. It stands out by demanding higher robustness out of the scene text understanding models against text of arbitrary shapes. Details about this competition and \textit{ArT} dataset can be found on the RRC competition website\footnote{\url{http://rrc.cvc.uab.es/?ch=14}}.
	
	The structure of this paper is as follows. Related work is presented in Sec. \ref{related} and details of the \textit{ArT} dataset are described in Sec. \ref{art_general}. Tasks that are involved in this competition can be found in Sec. \ref{task1}, \ref{task2}, \ref{task3} respectively with the task's description, evaluation metric and a brief discussion of participants' results in the subsections. This paper will then end with our conclusions in Sec. \ref{conclusions}.
	
	\section{Related Work}\label{related}
	Scene text reading methods have achieved significant progress alongside the evolution of the scene text benchmarks. The continuously emerging datasets follow several noticeable patterns: i) the size getting bigger, ii) the data becomes harder, and iii) the annotation becomes more flexible. In 2013, ICDAR2013 \cite{karatzas2013icdar} comprised 462 images with only well-focused rectangular-shaped text. On ICDAR2015 \cite{karatzas2015icdar} dataset, the number is increased to 1,500 and all the images were incidentally captured. Besides, the dataset introduces quadrilateral annotation to meet the variety of text shapes. In 2017, IC17-MLT \cite{nayef2017icdar2017} was introduced to challenge the community with the multi-script scene text reading problem in 9 different languages. Similarly, the size of the dataset increases to 18,000, and quadrangles were used as the ground truth format. 
	
	Recently, \cite{Chng2017TotalTextAC,yuliang2017detecting} pointed out although curved text instances are commonly found in the real world, they are rarely seen in the existing benchmarks. Besides, in the limited appearance of the curved text instances, their annotations are wildly loose with both the axis-aligned and quadrilateral bounding regions. Therefore, Total-text\cite{Chng2017TotalTextAC} and SCUT-CTW1500\cite{yuliang2017detecting} were collected with a great emphasis on curved text instance. Additionally, both of the datasets employed polygonal shape as the ground truth format for their annotations. These two benchmarks have quickly attracted the interests of the research community, motivating many promising text reading methods. Following the principles of both of the said datasets, the \textit{ArT} dataset aims to provide the community with a much larger data size to work with and a more comprehensive benchmark for future evaluations. 
	
	\section{The \textit{`ArT'} Dataset}\label{art_general}
	The dataset intended for this competition, \textit{ArT}, is a combination of Total-Text \cite{Chng2017TotalTextAC}\footnote{\url{https://github.com/cs-chan/Total-Text-Dataset}}, SCUT-CTW1500\cite{yuliang2017detecting}\footnote{\url{https://github.com/Yuliang-Liu/Curve-Text-Detector}}, Baidu Curved Text Dataset\footnote{A subset of LSVT} plus a large sample of newly collected images. The new images were collected following the same principles as \cite{Chng2017TotalTextAC,yuliang2017detecting}: i) At least one arbitrary-shaped text per image; ii) high diversity in terms of text orientations (i.e. large amounts of horizontal, multi-oriented, and curved text instances); iii) text instances are annotated with tight polygon ground truth format.
	
	\subsubsection{Type/source of images}
	Images in the \textit{ArT} dataset were collected via digital camera, mobile phone camera, Internet, Flickr, image libraries, and Google Open-Image \cite{krasin2016openimages}. Also, part of the new images that contain Chinese text are collected from Baidu Street View. Similar to most of the publicly scene text datasets, the images in \textit{ArT} contain scenes from both indoor and outdoor settings, with digitally born images included. Apart from the usual vision-related challenges (illumination, background complexity, perspective distortion, etc.), \textit{ArT} stands out in challenging scene text understanding models with the combination of different text orientations within one image.

	\subsubsection{Homogeneity of the dataset}
	The images from Total-Text \cite{Chng2017TotalTextAC}, SCUT-CTW1500 \cite{yuliang2017detecting} and Baidu Curved Text Dataset are similar in nature, they are i) from real world scenes, and ii) the images are mostly well focused. Hence, the combination is smooth in this aspect. However, since SCUT-CTW1500 considers Chinese script in their annotation while Total-Text does not; a refinement to the ground truth of Total-Text is done to annotate all the Chinese characters in it. In addition, the line-level annotation of the Latin scripts in SCUT-CTW1500 is also re-annotated to word-level.
	
	\subsubsection{Number of images}
	On top of the existing images (3055) from Total-Text \cite{Chng2017TotalTextAC} and SCUT-CTW1500 \cite{yuliang2017detecting}, 7111 images are added to make the \textit{ArT} dataset, one of the largest scene text datasets for arbitrary-shaped text. There is a total of 10,166 images in the \textit{ArT} dataset that is split into a training set with 5,603 images, and a testing set of 4,563 newly collected images. We acknowledge the Baidu team for annotating all the newly collected images via the Baidu crowd-sourcing platform.
	
	\subsubsection{Ground truth}
	It is worth pointing out that the polygon ground truth format employed in \textit{ArT} is different from all the previous RRC, which adopted the axis-aligned bounding box \cite{karatzas2013icdar,karatzas2015icdar}, or quadrilateral \cite{nayef2017icdar2017} as the ground truth format. Both of these annotation styles have two and four vertices respectively, which are intuitively inappropriate for the arbitrary-oriented Chinese and Latin text instances in \textit{ArT}, especially the curved text instances. Following the practice of the MLT dataset \cite{nayef2017icdar2017}, we annotated Chinese and Latin scripts at line-level and word-level granularities respectively. The transcription and the language type of annotated text instances are provided. Also, note that the coordinates of the polygon bounding boxes are labelled to have either 4, 8, 10, or 12 polygon vertices depending on their shape. All illegible text instances and symbols were labelled as ``Do Not Care'', which will not contribute to the evaluation result.
	
	\section{Organization}
	This competition is jointly organized by the University of Malaya, Malaysia; South China University of Technology, China; Baidu Inc, China; and the Computer Vision Centre (Autonomous University of Barcelona), Spain. There are monetary rewards to the winner of this challenge, which is sponsored by Baidu Inc.

	\section{Task 1: Scene Text Detection}\label{task1}
	\subsection{Description}
	The main objective of this task is to detect the location of every text instance in the input image. Given an input image, participants are expected to provide the spatial location and confidence score of each prediction.

	\subsection{Evaluation metrics}
	IoU-based evaluation protocol is adopted for this task by following \cite{yuliang2017detecting}. IoU (Intersection over Union) is a threshold-based evaluation protocol, with a default threshold of $0.5$. Results are reported both at $0.5$ and $0.7$ thresholds but only the H-Mean of the former threshold is used to determine the official ranking. To ensure fairness, the participants are required to submit confidence score for each detection, and thus all confidence thresholds are iterated to find the best H-Mean score. It is also worth mentioning that, \textit{ArT} will be the first RRC to handle unfixed detection output coordinates in Task 1 (Sec. \ref{task1}) and Task 3 (Sec. \ref{task3}).
	
	\subsection{Results and Discussion}
\begin{figure*}
		\begin{minipage}[t]{0.5\textwidth}
			\centering
			\includegraphics[keepaspectratio=true, scale=0.175]{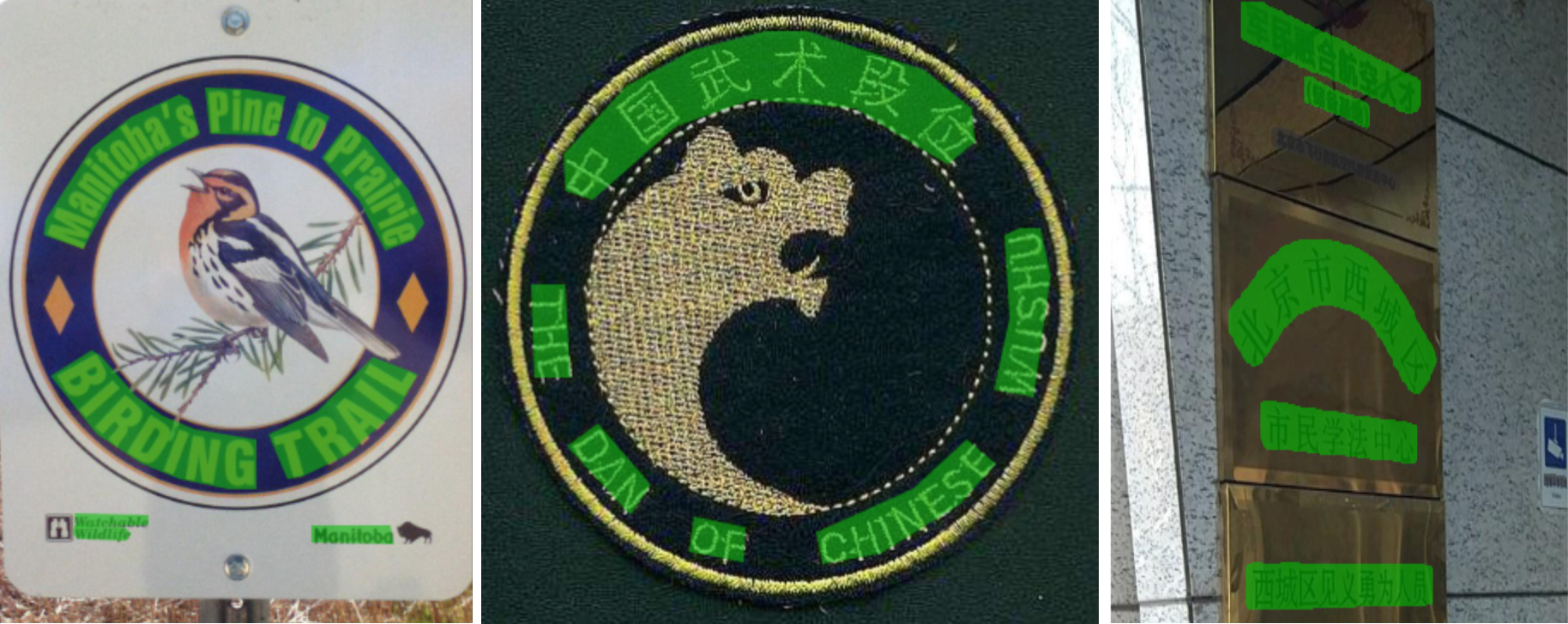}
			\subcaption{ }\label{fig:pil_success}
		\end{minipage}
		\hfill
		\begin{minipage}[t]{0.5\textwidth}
			\includegraphics[keepaspectratio=true, scale=0.2]{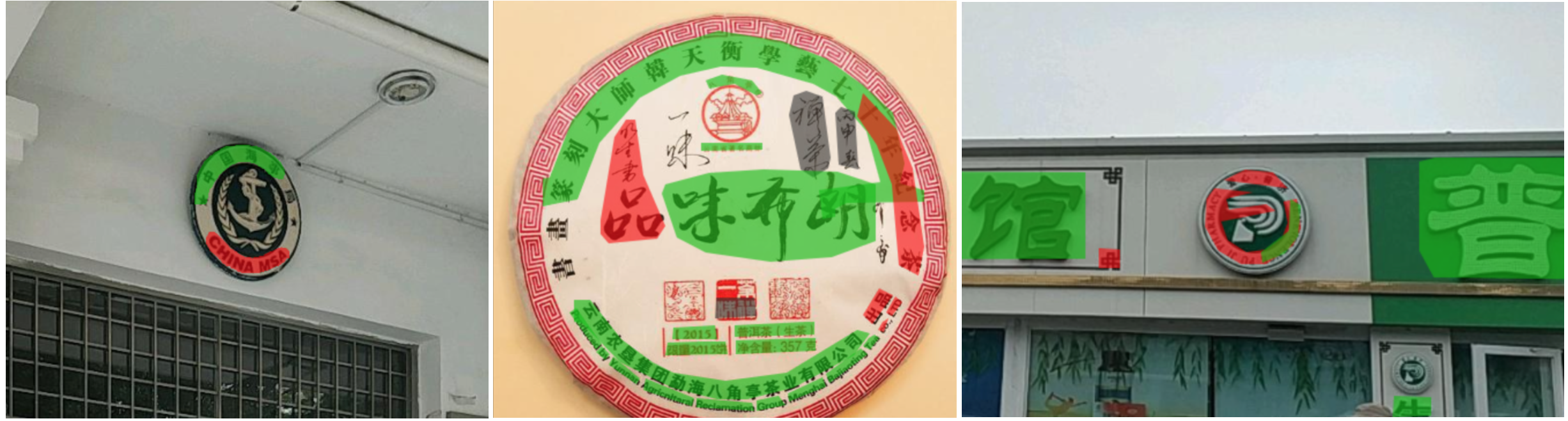}
			\subcaption{ }\label{fig:pil_failure}
		\end{minipage}
		\caption{Successful (a) and failure (b) detection examples of \textbf{\textit{Pil-Mask-RCNN}}. \textbf{Green}: True Positive, \textbf{Red}: False Positive and False Negative.}\vspace{-.1in}
		\label{fig:pil_suc_and_fail}
	\end{figure*}

	For Task 1, we received 48 submissions with 35 of them submitted from unique participants. The average H-mean score for Task 1 is 67.46\%. The first place of this task is \textbf{\textit{Pil-Mask-RCNN}} by Wang \etal from Institute of Computational Technology, CAS, China, with the winning H-mean score of 82.65\%. The proposed method is built based on the Mask R-CNN pipeline with two different backbone networks: Senet-152 and Shuffle-net v2. Figure \ref{fig:pil_success} illustrates some of the successful examples. The visualization of its results show that the detection regions are of high quality: smooth and tight. Besides, it appears to be robust against the language variant of the text instances as well (i.e. Chinese and Latin scripts). We also investigated the failure examples of the winning method (as seen in Figure \ref{fig:pil_failure}), the common problems are: i) under segmenting (combining multiple text instances into one), ii) mistaken group of crowded text instances in a small area (especially Chinese characters), and iii) small text instances. We notice that most of the top performing methods (both runners-up included) are based on the Mask-RCNN pipeline. Also, most of the participants (except 4 submissions) design their models to produce polygon bounding region as the detection output, which align with the emphasis of this competition - tightness of detection outputs. 
	
\begin{figure*}[htbp]
		\centering
		\begin{minipage}[t]{0.3\textwidth}
			\includegraphics[keepaspectratio=true, scale=0.22]{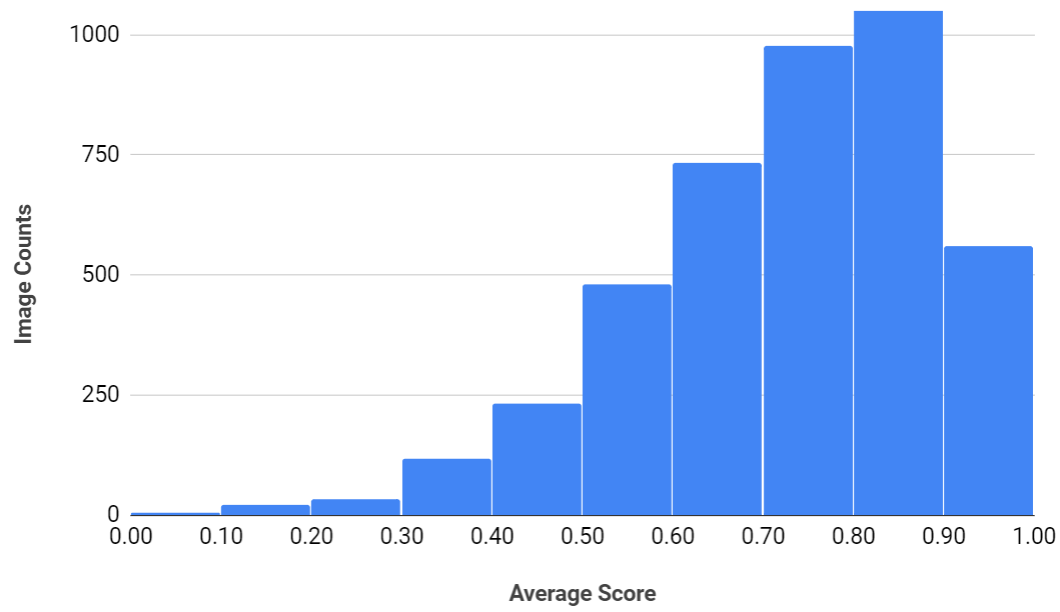}
			\subcaption{Task 1}\label{subfig:t1}
		\end{minipage}
		\hfill
		\begin{minipage}[t]{0.3\textwidth}
			\centering
			\includegraphics[keepaspectratio=true, scale=0.22]{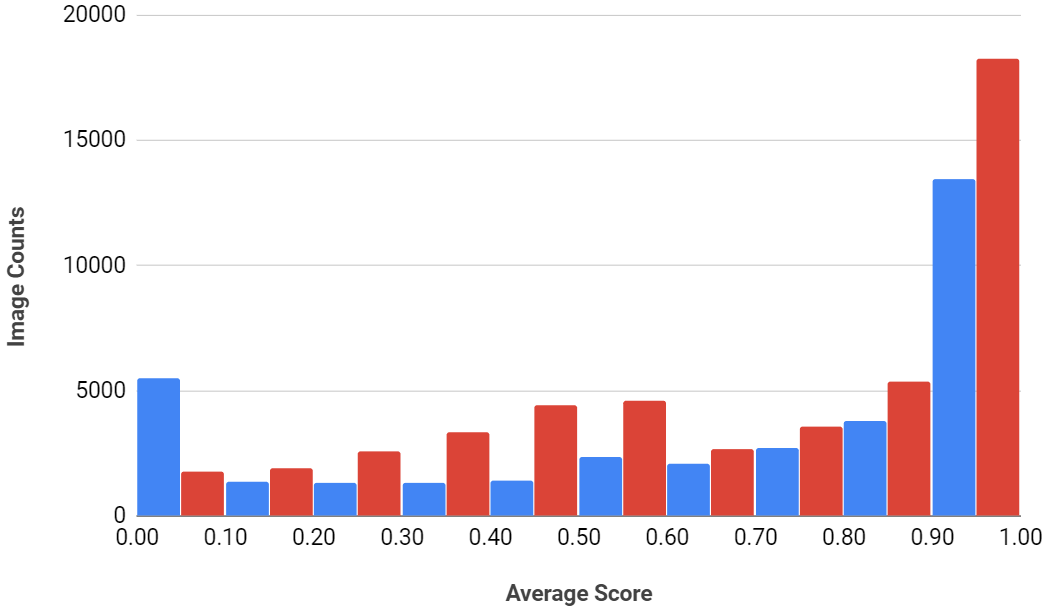}
			\subcaption{Task 2.1 (blue) and Task 2.2 (red)}\label{subfig:t2}
		\end{minipage}
		\hfill
		\begin{minipage}[t]{0.3\textwidth}
			\centering
			\includegraphics[keepaspectratio=true, scale=0.22]{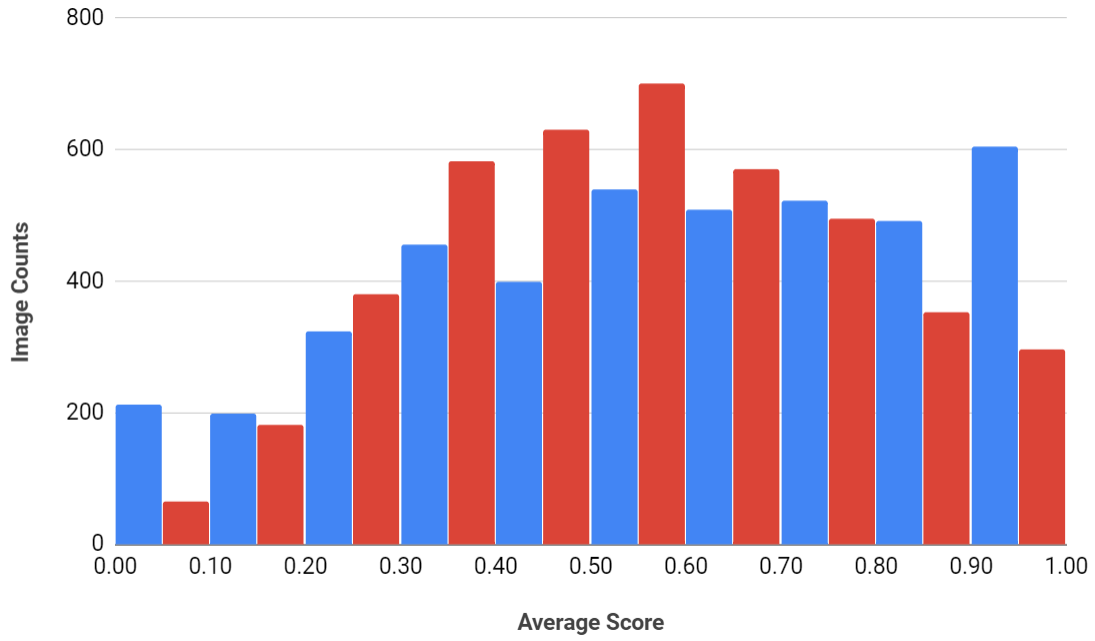}
			\subcaption{Task 3.1 (blue) and Task 3.2 (red)}
			\label{subfig:t3}
		\end{minipage}
		\caption{Histogram of the average score by all submissions of each test set image.} 
		\label{fig:T1_hist}
	\end{figure*}

\begin{figure*}[!]
		\begin{minipage}[t]{0.24\textwidth}
			\includegraphics[keepaspectratio=true, height=1.0\textwidth, width=1.0\textwidth]{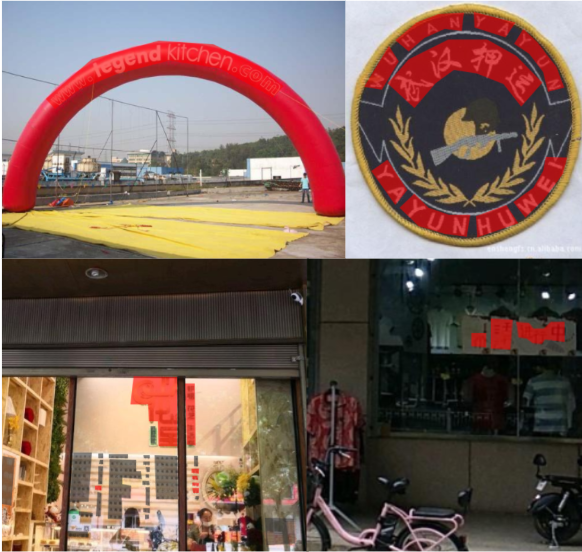}
			\subcaption{Task 1}\label{subfig:lh_t1}
		\end{minipage}
		\begin{minipage}[b]{0.24\textwidth}
			\begin{minipage}[t]{1.0\textwidth}
				\includegraphics[keepaspectratio=true, width=1.0\textwidth]{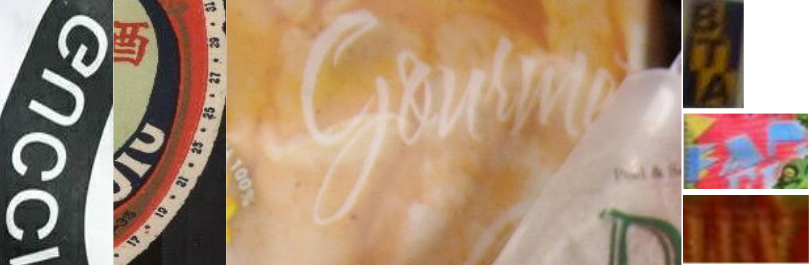}
				\subcaption{Task 2.1}\label{subfig:lh_t2_1}
			\end{minipage}
			\begin{minipage}[t]{1.0\textwidth}
				\includegraphics[keepaspectratio=true, width=1.0\textwidth]{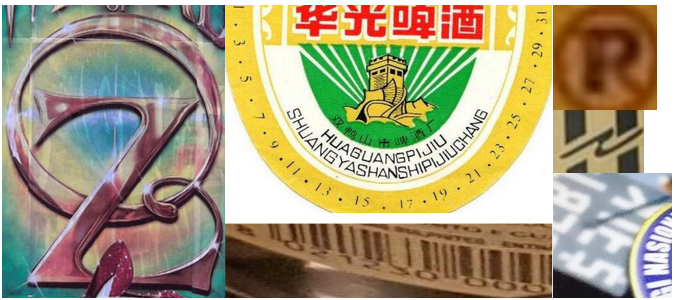}
				\subcaption{Task 2.2}\label{subfig:lh_t2_2}
			\end{minipage}
		\end{minipage}
		\begin{minipage}[t]{0.24\textwidth}
			\includegraphics[keepaspectratio=true, height=1.0\textwidth, width=1.0\textwidth]{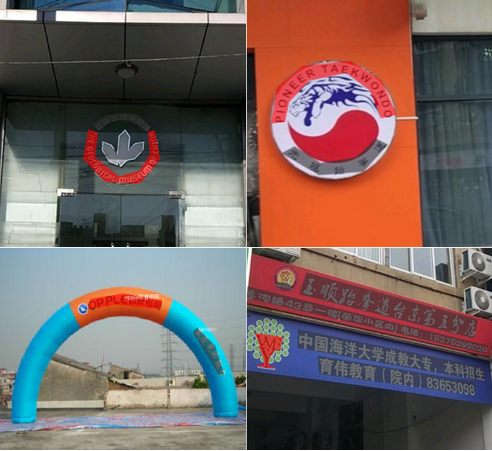}
			\subcaption{Task 3.1}\label{subfig:lh_t3_1}
		\end{minipage}
		\begin{minipage}[t]{0.24\textwidth}
			\includegraphics[keepaspectratio=true, height=1.0\textwidth, width=1.0\textwidth]{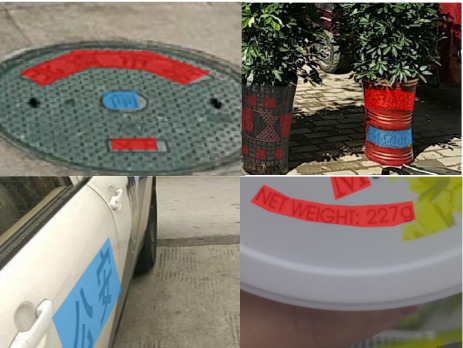}
			\subcaption{Task 3.2}\label{subfig:lh_t3_2}
		\end{minipage}
		\caption{Example images with low average H-mean score (i.e. 0-0.1). \textbf{Red}: Misdetections, \textbf{Blue}: False recognitions.
		} \vspace{-.1in}
		\label{fig:T1_hmean_low}
	\end{figure*}

	The ranking of Task 1 is tabulated in Table \ref{tab:Task1}. Note that the top 3 teams between 0.5 IoU and 0.7 IoU are different - the original runner up - \textit{\textbf{NJU-ImagineLab}} is overtook by \textit{\textbf{ArtDet-v2}} and drops to the fourth place. Meanwhile, Figure \ref{subfig:t1} is the histogram of the average H-mean scores of each image in the testing set. As we can see, most images have the average H-mean scores between 0.8 to 0.9, followed by 0.7-0.8 so forth. The challenging images with 0 - 0.1 H-mean score can be seen in Figure \ref{fig:T1_hmean_low}.

	\begin{table}
		\centering
		\begin{adjustbox}{width=0.98\linewidth, height=2.3in}
			\begin{tabular}{lccccc}
				\toprule
				Task & Rank &Method Name & Affiliation & H-mean (IoU $>$ 0.5) & H-mean (IoU $>$ 0.7) \\
				\midrule
				1 & 1 &pil\_maskrcnn & Institute of Computing Technology, Chinese Academy of Sciences & 82.65 & 76.06 \\
				& 2 &NJU-ImagineLab & Nanjing University & 80.24 &  	70.33 \\
				& 3 &ArtDet-v2 & Sogou-OCR team & 79.48 & 72.01 \\
				& 4 &baseline\_polygon& Beihang University & 78.79 &  68.36 \\
				& 5 &CUTeOCR & CUHK, HIT & 78.36 & 71.31 \\
				& 6 &Sg\_ptd& Sogou Tech & 77.42 &  65.04 \\
				& 7 &Alibaba-PAI& Alibaba Group & 76.1 & 64.41 \\
				& 8 &Fudan-Supremind Detection v3 & Fudan University & 75.24 & 64.76 \\
				& 9 &SRCB\_Art& SRCB & 75.02 & 65.25 \\
				& 10 &A scene text detection method based on maskrcnn & Fudan University & 74.72 & 65.24 \\
				& 11 &DMText\_art& Tencent & 74.43 & 65.94 \\
				& 12 &TEXT\_SNIPER& IIE, CAS & 73.74 & 59.36 \\
				& 13 &CLTDR& Chinese Academy of Sciences & 73.32 & 64.66 \\
				& 14 &CRAFT& Clova AI OCR Team, NAVER/LINE Corp & 72.85 &  56.16 \\
				& 15 &Sogou\_MM& Sogou Inc Sogou\_MM team & 72.69 & 60.61 \\
				& 16 &QAQ& Institute of Automation, Chinese Academy of Sciences & 72.21 & 55.60 \\
				& 17 &MaskDet & MetaSota.ai & 71.44 & 59.07 \\
				& 18 &fdu\_ai & Fairleigh Dickinson University & 70.4 &  61.11 \\
				& 19 &CCISTD & Peking University & 69.47 & 61.09 \\
				& 20 &Mask RCNN& - & 68.95 & 59.07 \\
				& 21 &TextMask\_V1 & - & 68.92 & 60.63 \\
				& 22 &MFTD: Mask Filters for Text Detection & - & 67.27 & 55.92 \\
				& 23 &Art detect by vivo & VIVO AI Lab & 66.92 & 55.55 \\
				& 24 &PAT-S.Y& - & 66.72 & 54.22 \\
				& 25 &DMCA & IIE,CAS & 66.45 &  52.25 \\
				& 26 &TMIS & USTC-iFLYTEK & 66.01 & 56.53 \\
				& 27 &mask rcnn& - & 63.81 &  50.11 \\
				& 28 &Unicamp-SRBR-PN-1& SRBR, Unicamp & 62.37 & 46.46 \\
				& 29 &TP & Shanghai Jiao Tong University & 62.18 &  50.86 \\
				& 30 &Improved Progressive scale expansion Net& - &61.88 & 49.50 \\
				& 31 &1& - & 58.2 & 41.66 \\
				& 32 &TextCohesion\_1  & Zhengzhou University & 53.2 &  42.40 \\
				& 33 &EM-DATA& - & 51.99 & 32.22 \\
				& 34 &RAST: Robust Arbitrary Shape Text Detector& - & 47.3 & 36.51 \\
				& 35 &MSR & - & 0.50 &  0.07 \\ 
				\midrule
				Task & Rank & Method Name & Affiliation & Accuracy & 1-N.E.D\\
				\midrule
				2.1 & 1 &PKU\_Team\_Zero & MEGVII (Face++), Peking University & 74.30 & - \\
				& 2 &CUTeOCR & CUHK, HIT & 73.91 &  - \\
				& 3 &CRAFT (Preprocessing) + TPS-ResNet & Clova AI OCR Team, NAVER/LINE Corp & 73.87 & - \\
				& 4 &NPU-ASGO & Northwestern Polytechnical University & 71.82 & - \\
				& 5 &CIGIT and XJTLU & CIGIT, XJTLU & 70.73 & - \\
				& 6 &Attention based method for scene text recognition & SenseTime Group & 70.39 &  - \\
				& 7 &Ensemble and post processes & - & 69.15 &  - \\
				& 8 &CSN-ED & USTC-iFLYTEK & 67.32 &  - \\
				& 9 &Alchera AI & Alchera AI & 66.81 & - \\
				& 10 &Irregular Text Recognizer with Attention Mechanism & Pennsylvania State University & 64.45 &  - \\
				& 11 &class\_5435\_rotate &  Beihang University & 63.86 & - \\
				& 12 &MatchCRNN & MetaSota.ai &  58.03 &  - \\
				& 13 &Arbitrary shape scene text recognition based on CNN and Attention Enhanced Bi-directional LSTM & - & 56.09 & - \\
				& 14 &Fudan-Supremind Recognition & Fudan University & 50.56 &  - \\
				& 15 &LCT\_OCR & IIE, CAS & 47.31 & - \\
				& 16 &So Cold 2.0 & - & 45.30 &  - \\
				& 17 &task2x & - & 38.08 & - \\
				\midrule
				2.2 & 1 &CRAFT (Preprocessing) + TPS-ResNet & Clova AI OCR Team, NAVER/LINE Corp & - & 85.32 \\
				& 2 &Attention based method for arbitrary-shaped scene text recognition & SenseTime Group & - & 	85.20 \\
				& 3 &CSN-ED & USTC-iFLYTEK & - &  81.23 \\
				& 4 &class\_5435\_rotate &  Beihang University & - & 80.60 \\
				& 5 &MatchCRNN & MetaSota.ai & - & 72.61 \\
				& 6 &Ensemble and post processes & - & - &  71.27 \\
				& 7 &So Cold 2.0 & - & - &  69.76 \\
				& 8 &Fudan-Supremind Recognition & Fudan University & - &  66.15 \\
				& 9 &CUTeOCR & CUHK, HIT & - &  65.38 \\
				& 10 &PKU\_Team\_Zero & MEGVII (Face++), Peking University & - & 65.06 \\
				& 11 &NPU-ASGO & Northwestern Polytechnical University & - & 63.82 \\
				& 12 &CIGIT and XJTLU & CIGIT and XJTLU & - & 63.15 \\
				& 13 &Alchera AI & Alchera AI & - & 	61.61 \\
				& 14 &Irregular Text Recognizer with Attention Mechanism & Pennsylvania State University & - & 61.42 \\
				& 15 &LCT\_OCR & IIE, CAS & - & 59.77 \\
				& 16 &task2x & - & - & 56.53 \\
				& 17 &Arbitrary shape scene text recognition based on CNN and Attention Enhanced Bi-directional LSTM & - & - & 54.49 \\
				\midrule
				Task & Rank & Method Name & Affiliation & Accuracy H-mean & 1-N.E.D\\
				\midrule
				3.1 & 1 &baseline\_0.5\_class\_5435 & Beihang University & 52.45 & 53.86 \\
				& 2 &Alibaba-PAI & Alibaba Group & 57.32 &  53.36\\
				& 3 &QAQ3 &  Institute of Automation, Chinese Academy of Sciences & 45.57 &  46.01\\
				& 4 &Detection-Recognition & USTC-iFLYTEK & 48.64& 45.84\\
				& 5 &CLTDR & Chinese Academy of Sciences & 44.71 & 44.49\\
				& 6 &So Cold 2.0 & - & 37.09&  39.71\\
				& 7 &task3 & - & 37.48& 34.03\\
				& 8 &CRAFT + TPS-ResNet v1& Clova AI OCR Team, NAVER/LINE Corp & 31.68 & 27.21 \\
				\midrule
				3.2 & 1 &baseline\_0.5\_class\_5435 & Beihang University & 50.17 & 54.91 \\
				& 2 &Alibaba-PAI & Alibaba Group & 53.48 &  51.68 \\
				& 3 &QAQ3 &  Institute of Automation, Chinese Academy of Sciences & 47.48 & 49.10 \\
				& 4 &CLTDR & Chinese Academy of Sciences & 45.65 & 48.78 \\
				& 5 &Detection-Recognition & USTC-iFLYTEK & 46.13 & 48.03 \\
				& 6 &So Cold 2.0 & - & 34.14 &  39.58 \\
				& 7 &task3 & - & 38.58 & 37.65 \\
				& 8 &CRAFT + TPS-ResNet v1& Clova AI OCR Team, NAVER/LINE Corp & 32.26 &  29.58 \\
				\bottomrule
			\end{tabular}
		\end{adjustbox}
		\caption{Official ranking of all the tasks in the RRC-ArT competition. Zoom in for a better view.} \vspace{-10pt} \label{tab:Task1}
	\end{table}

	\section{Task 2: Scene Text Recognition}\label{task2}
	\subsection{Description}
	The main objective of this task is to recognize every character in a cropped image patch. The input patterns of this task are the cropped image patches with corresponding text instances, and the relative polygon spatial coordinates. Participants are asked to provide the recognized string of characters as output. Nevertheless, it is up to participants to choose if they want to utilise the polygon coordinates as they are provided as optional information. Furthermore, we decided to further break down Task 2 into two subcategories: i) Task 2.1 - Latin script only recognition, and ii) Task 2.2 - Latin and Chinese scripts recognition. We hope that such a split could make this task friendlier for non-Chinese participants, as the aim of this competition is to detect and recognize arbitrary-shaped text. Participants are required to make a single submission only regardless of the scripts. We evaluated all submissions under two categories, Latin and mixed (Latin and Chinese) scripts. When evaluating the recognition performance for Latin script, all non-Latin transcriptions will be treated as ``Do Not Care'' regions.
	
	\subsection{Evaluation metrics}
	For Task 2.1, case-insensitive word accuracy is used as the primary challenge metric. Apart from this, all the standard practices for text recognition evaluation are followed. For example, symbols in the middle of ground truth text instances are considered but symbols such as ( !?.:,*"()·[]/'\_ ) at the beginning and at the end of both the ground truth and the submissions are removed. For Task 2.2, the Normalized Edit Distance metric (1-N.E.D specifically, which is also used in the ICDAR 2017 competition, RCTW-17 \cite{rctw}) are treated as the ranking metric. The reason of utilizing 1-N.E.D as the official ranking metric for Task 2.2 is motivated by the fact that Chinese scripts usually contain more characters than the Latin scripts, which makes word accuracy metric too harsh to evaluate Task 2.2 fairly. In the 1-N.E.D evaluation protocol, all characters (Latin and Chinese) will be treated in a consistent manner. To avoid ambiguities in the annotations, we performed several pre-processing steps before the evaluation process: 1) English letters are treated as case insensitive; 2) Chinese traditional and simplified characters are treated as the same label; 3) Blank spaces and symbols will be removed; 4) All illegible images will not contribute to the evaluation result.

	\subsection{Results and Discussion}
\begin{figure}
		\begin{minipage}[t]{0.48\textwidth}
			\centering
			\includegraphics[keepaspectratio=true, scale=0.26]{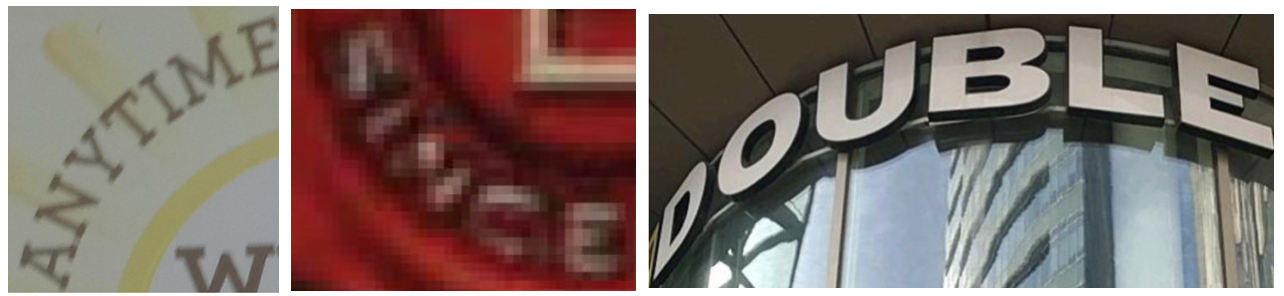}
			\subcaption{ }\label{fig:T2_1_PKU_success}
		\end{minipage}
		\vspace{-.1in}
		\begin{minipage}[t]{0.48\textwidth}
			\centering
			\includegraphics[keepaspectratio=true, scale=0.25]{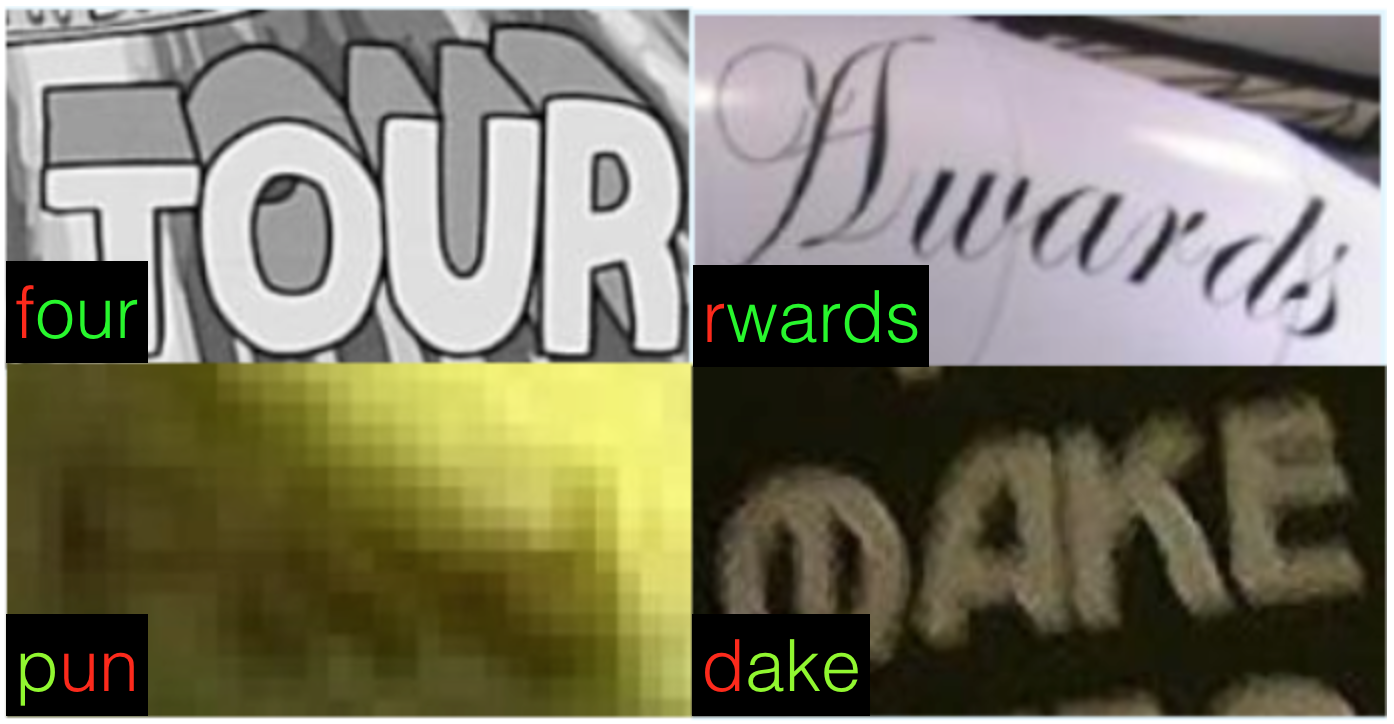}
			\subcaption{ }\label{fig:T2_1_PKU_failure}
		\end{minipage}
		\caption{Successful (a) and failure (b) recognition examples of \textbf{\textit{PKU\_Team\_Zero}}.} \vspace{-.1in}
		\label{fig:pku_suc_and_fail}
	\end{figure}

	For Task 2, there are 22 unique submissions from 17 unique teams. Starting with Task 2.1, the average accuracy score of this task is 62.47\%. The winner of this task is \textbf{\textit{PKU\_Team\_Zero}} by Shangbang \etal from MEGVII (Face++) and Peking University, China, with the winning score of 74.30\%. It comprises of three major modules: 1) A detection module that can provide the spatial coordinates of the text (in polygon vertices) within the cropped image; 2) a spatial transformer that can straighten the image based on the coordinates; and 3) an attention RNN model for recognizing words. 
	We notice that all three winning models have similar pipelines - all of them rectify the cropped image patches (i.e. straighten the text region, in turn removing background) before recognizing the word in it. This shows that the polygon ground truth format instead of the normal bounding box is indeed crucial in the problem of recognizing curved or any arbitrary text instances. Besides, another similarity is that all three of them employ attention mechanism in their RNN word recognition module. Qualitative results of the \textbf{\textit{PKU\_Team\_Zero}} method can be seen in Figure \ref{fig:T2_1_PKU_success}. The method has demonstrated its outstanding ability in recognizing curved text instances of challenging attributes in real world scene. On the other hand, Figure \ref{fig:T2_1_PKU_failure} illustrates some of the failure examples. The failure cases are mainly caused by unusual font types and severely blurred patch. 
	
\begin{figure}[t]
		\begin{minipage}[t]{0.48\textwidth}
			\centering
			\includegraphics[keepaspectratio=true, scale=0.34]{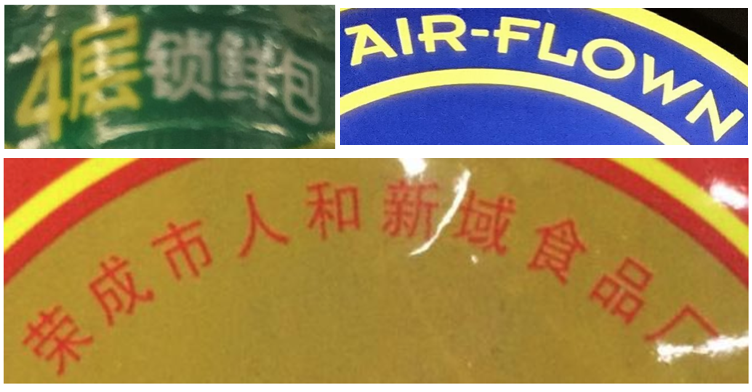}
			\subcaption{ }\label{fig:T2_2_craft_success}
		\end{minipage}
		\vspace{-.1in}
		\begin{minipage}[t]{0.48\textwidth}
			\centering
			\includegraphics[keepaspectratio=true, scale=0.28]{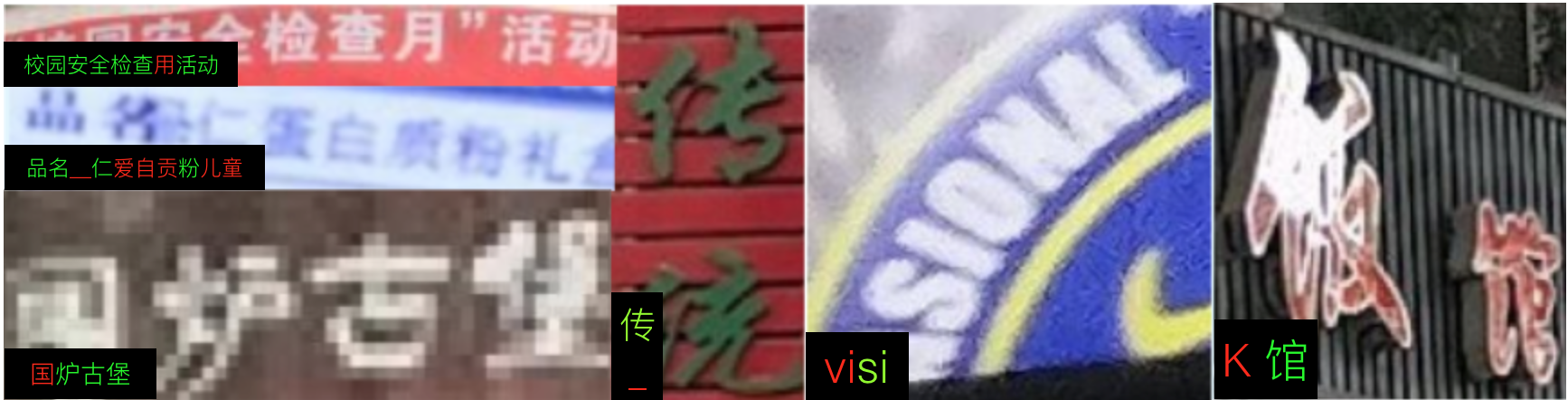}
			\subcaption{ }\label{fig:T2_2_craft_failure}
		\end{minipage}
		\caption{Successful (a) and failure (b) recognition examples of \textbf{\textit{CRAFT (Preprocessing) + TPS-ResNet}}.}
		\label{fig:craft_suc_and_fail}
	\end{figure}

	The top three methods for Task 2.2 are quite different from Task 2.1. The average 1-N.E.D score of this sub-task is 68.43\%, and the winner of this task is \textbf{\textit{CRAFT (Preprocessing) + TPS-ResNet}} by Baek \etal from Naver Corporation which scores 85.32\%. This method also has three major modules: detection, rectification, and recognition. Specifically, it adopts CRAFT \cite{baek2019character} as its text detector, Thin-Plate-Spline (TPS) based Spatial Transformer Network as its image normalizer, and a BiLSTM with attention as its text recognizer. Figure \ref{fig:T2_2_craft_success} shows some successful examples of the said method, it appears that the method is robust against curved text instances on both the Chinese and Latin scripts. Failure cases can be seen in Figure \ref{fig:T2_2_craft_failure}, where it fails in 1) Chinese character with similar appearance, 2) vertical oriented text, 3) blurred patch, and 4) interestingly Chinese character that looks like `K' under perspective distortion and illumination. 
	
	The global performance of Task 2 is summarized in Figure \ref{subfig:t2}. From this figure, we notice two obvious spikes in the 0-0.1 and 0.9-1.0 bars for Task 2.1 (blue). This phenomenon is because of the attribute of accuracy scoring mechanism (i.e. 1 for getting every character recognized and 0 otherwise). Meanwhile, in Task 2.2 (red), we see a smoother distribution between 0 to 1. As we can see, most of the patches have a high average 1-N.E.D score (between 0.9 and 1).

	\section{Task 3: Scene Text Spotting}\label{task3}
	\subsection{Description}
	The main objective of this task is to detect and recognize every text instance in the provided image in an end-to-end manner. Given an input image, the output must be the spatial location of every text instance at word-level for Latin script and line-level for Chinese script together with the predicted word for each detection. Similar to RRC 2017 \cite{nayef2017icdar2017}, a generic vocabulary list (90K common English words) will be provided as a reference for this task. Identical to Task 2, we break Task 3 down into two subcategories: i) Task 3.1 Latin script only text spotting, and ii) Task 3.2 Latin and Chinese scripts text spotting.
	
	\subsection{Evaluation metrics}
	For Task 3, we first evaluate the detection result by calculating its IoU with the corresponding ground truth. Detection regions with an IoU value higher than 0.5 are then matched with the recognition ground truth (i.e. the transcript ground truth of that particular text region). Meanwhile, in the case of multiple matches, we only consider the detection region with the highest IOU, the rest of matches will be counted as False Positive. The pre-processing steps for the recognition part are the same as Task 2 and all Chinese text regions are ignored in Task 3.1. Also, it is worth mentioning that although the results of case-insensitive word accuracy H-mean and 1-N.E.D will be reported but the official ranking metric for both sub-tasks are 1-N.E.D.
	
	\subsection{Results and Discussion}
\begin{figure}[!]
		\begin{center}
			\includegraphics[keepaspectratio=true, height=0.4\linewidth, width=\linewidth]{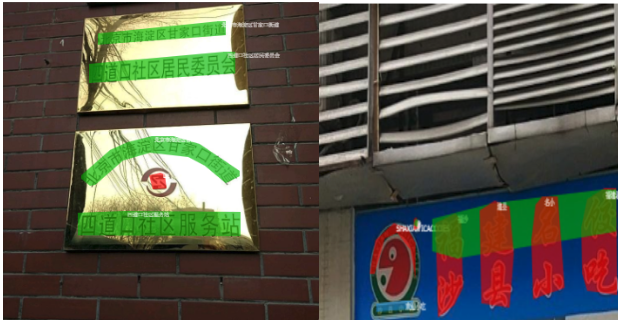}
		\end{center}
		\caption{Successful \textbf{(left)} and failure \textbf{(right)} examples of \textbf{\textit{baseline\_0.5\_class\_5435}}.} \vspace{-.1in}
		\label{fig:T3_baseline_0.5}
	\end{figure}

	Task 3 received 8 submissions from 8 individual teams. It is also the hardest task among all tasks - the average accuracy H-mean score for Task 3.1 is only 44.37\%. The method that ranks the first place is \textbf{\textit{baseline\_0.5\_class\_5435}} by Jinjin Zhang from Beihang University, China with the accuracy H-mean score of 52.45\%. The winning method has a segmentation-based detector and an attention-based recognizer. Zhang mentioned that the method is modelled to have 5,435 classes for the recognition task. Besides, extra training data from LSVT, ICDAR2017, COCO-Text, ReCTS, and augmented data were used to train their recognition network. The top three winners of Task 3.2 is the same as Task 3.1, with a slightly higher average 1-N.E.D score - 44.91\%. Figure \ref{fig:T3_baseline_0.5} depicts several successful and failure examples. As observed, in a high contrast setting (left figure), every text instance is well detected and recognized by the model; while the challenging example on the right confused the method with multiple possible combinations of the text instances. To be specific, the four vertical red regions are evaluated as false positives; the actual ground truths are supposed to be two text regions arranged from left to right (top and bottom), making up two Chinese words. Such an example could potentially be solved by instilling semantic information (e.g. the specific language knowledge) into the text spotting model.

	In contrast to Task 1 and 2, the histogram of Task 3 (in Figure \ref{subfig:t3}) demonstrates the most distributed pattern across the score range. Note that for Task 3.1 (blue) the spike at the 0.9-1.0 (in contrast to the low count of Task 3.2 (red)) is due to the fact that Chinese scripts are counted as ``Do Not Care'' regions, which makes it easier to score a full mark on Chinese text dominant images. In general, most of the images have 0.4 to 0.6 average score which reflect the challenging aspect of this task. Again, Figure \ref{subfig:lh_t3_1} and \ref{subfig:lh_t3_2} shows some of the most challenging images in the test set with 0 to 0.1 average score.

	\section{Conclusions}\label{conclusions}
	The ICDAR2019 Robust Reading Challenge on \textit{ArT} received an overwhelming number of submissions, which is a delightful outcome considering that scene text understanding works with curved text in consideration were rarely seen before the introduction of the Total-Text and SCUT-CTW1500 datasets recently. Although the scene text understanding community has seen tremendous improvements in very recent years, the gap between the research-end and the application-end still exists. The main motivation behind \textit{ArT} dataset and this challenge is to encourage both the academic and industrial fields to look into the arbitrary orientation or shape aspect of text instances in the wild. 
	
	The score of the top three winners in all tasks are close to each other, which is a good indication of where the state-of-the-arts resides at the moment. By taking a deeper look into the submission models, segmentation based methods seem to dominate the arbitrary-shaped text detection. Besides, we also find that the current IoU metric has many drawbacks; for example, some of the detections that miss several characters are still being rewarded with 100\% recall. Therefore, a better and more reasonable metric such as the recent TIoU metric \cite{liu2019tightness} may be worth practising in the future. In the recognition tasks, popular and high performing models share similar pipelines, which includes rectifying the text patches before recognizing them with an attention RNN/LSTM module. To this end, text spotting seems to be the most challenging task with the lowest winning H-mean score.

	\bibliographystyle{IEEEtran}
	\bibliography{egbib}

\end{document}